# Çoklu Biyometrik Sistem ile İkiz Tanıma: Ses ve Kulak Tabanlı Çoklu Model ile El Yapımı ve Derin Öğrenme Tabanlı Çoklu Algoritma

# Twins Recognition with Multi Biometric System: Handcrafted-Deep Learning Based Multi Algorithm with Voice-Ear Recognition Based Multi Modal

*Yazarlar Gizlenmiştir*

*Özetçe*—Teknolojinin gelişmesi ile birlikte biyometrik sistemlerin kullanım alanları ve önemi artmaya başlamıştır.Her insanın karakteristik özellikleri birbirinden farklı olduğu için tek modelli bir biyometrik sistem başarılı sonuçlar verebilmektedir. Ancak ikizi olan insanların karakteristik özellikleri birbirine çok yakın olmasından dolayı, bireylerin birden çok karakteristik özelliğini içeren çoklu biyometrik sistem kullanımı daha uygun olacak ve tanıma oranını artıracaktır. Bu çalışmada insanları diğer insanlardan ve kendi ikizlerinden ayırt etmek için çoklu algoritma ve çoklu model birleşiminden oluşan çoklu biyometrik bir tanıma sistemi geliştirilmiştir. Çoklu model için kulak ve ses biyometrik verileri kullanılmış olup, veri setinde 38 çift ikizin kulak görüntüleri ve ses kayıtları bulunmaktadır. Klasik(el yapımı) ve derin öğrenme algoritmaları kullanarak ses ve kulak tanıma oranları elde edilmiştir. Elde edilen sonuçlar skor seviyesi füzyon yöntemi ile birleştirilerek sıra-1'de %94,74 ve sıra-2'de %100 başarıya ulaşılmıştır.

*Anahtar Kelimeler* — *çoklu biyometrik sistem; sestanıma; kulak tanıma; derin öğrenme;skor seviyesi füzyon.*

*Abstract*— With the development of technology, the usage areas and importance of biometric systems have started to increase. Since the characteristics of each person are different from each other, a single model biometric system can yield successful results. However, because the characteristics of twin people are very close to each other, multiple biometric systems including multiple characteristics of individuals will be more appropriate and will increase the recognition rate. In this study, a multiple biometric recognition system consisting of a combination of multiple algorithms and multiple models was developed to distinguish people from other people and their twins. Ear and voice biometric data were used for the multimodal model and 38 pair of twin ear images and sound recordings were used in the data set. Sound and ear recognition rates were obtained using classical (hand-crafted) and deep learning algorithms. The results obtained were combined with the score level fusion method to achieve a success rate of 94.74% in rank-1 and 100% in rank -2.

I. GİRİŞ

İkiz nüfusunun artması ile birlikte biyometrik sistemlerin ikizlerin olduğu veri setlerinde de başarılı halde çalışması gerekmektedir. Bu kapsamda ikizler ve çoklumodelbiyometrik sistemler üzerine çalışmalar yapılmaktadır:

İnsanların yüzündeki belirli izleri kullanarak ikizleri ayırma üzerine çalışma yapılmıştır. Bunun için yüzdeki ben, çil, kalıcı leke, kalıcı yara izi, kalıcı sivilce ve doğum lekesi gibi izler tanıma için kullanılmıştır. 89 çift ikizden oluşan 178 kişi ve onların 477 adetgörüntüsü veri seti olarak kullanılmıştır [1].

Yüz tanıma algoritmaları kullanarak ikiz olan iki kişiyi birbirinden ayırt etmek için çalışmalar yapılmıştır [2, 3].

Lakshmi Priya ve Pushpa Raniparmak izi, yüz görüntüleri ve dudak izi görüntüleri kullanılarak çoklu model bir biyometrik sistem tasarlamıştır. Veri setinde 214 çift ikiz bulunmaktadır [4].

Bayan Omar Mohammed ve Siti Mariyam Shamsuddin parmak izi ve el yazısını kullanarak çoklu model bir biyometrik tanıma sistemi gerçekleştirmiştir. Veri setinde 20 çift ikiz bulunmaktadır [5, 6].

Çoklu model derin öğrenme üzerine yapılan bir çalışmada çapraz modelli öğrenme ve paylaşılan gösterimli öğrenme algoritmaları gerçekleştirilmiştir. Bu çalışmada CUAVE ve AVLetters veri setleri kullanılmıştır [7].

## II. ÇOKLU MODEL ÇOKLU ALGORİTMA

Çalışmada kullanılan ikizler veri seti 39 çift ikizin ses kayıtlarından ve sağ-sol kulak görüntülerinden oluşmaktadır. Ancak, veri setinde 1 kişinin sağ kulak görüntüsü olmadığı için bu çalışmada 38 çift ikizin ses kaydı ve kulak görüntüleri kullanılmıştır. Bir kişi için 3 ses kaydı bulunmaktadır. Ses kayıtlarının 2 tanesi eğitim seti için kullanılmış olup 1 tanesi de test seti için kullanılmıştır. Her bir kişi için birer adet sol ve sağdan çekilmiş olan toplam 2 adet kulak görüntüsü bulunmaktadır. Sol kulak görüntüsü eğitim seti için sağ kulak görüntüsü de test seti için kullanılmıştır.

## III. SES TANIMA ALGORİTMALARI

### A. Mel Frekans Kepstral Katsayıları

Mel Frekans Kepstral Katsayıları(MFKK) ses tanımada en çok kullanılan öznitelik çıkarma algoritmalarından birisidir. İnsanın duyma algısına göre modellenmiştir. Ses sinyaline ilk önce bir ön-filtreleme uygulanır. Filtrelenmiş sinyal çerçevelere bölünür ve her çerçeveye "hamming penceresi", ardından bu değerlere hızlı fourier dönüşümü uygulanır ve bu değerler de "mel" filtresinden geçirilir. Filtrelenen değerlerin logaritması alınır ve ardından ayrık kosinüs dönüşümü işlemi ile MFKK elde edilir.

### B. Yapay Sinir Ağları: Uzun Kısa-Vadeli Bellek Algoritması

Uzun Kısa-Vadeli Bellek(UKVB), uzun vadeli bağımlılık sorununu çözmek için geliştirilmiş özel bir Tekrarlayan Yapay Sinir Ağı(TYSA) türüdür. Bu yöntem 1997 yılında Hochreiter ve Schmidhuber tarafından önerilmiştir.

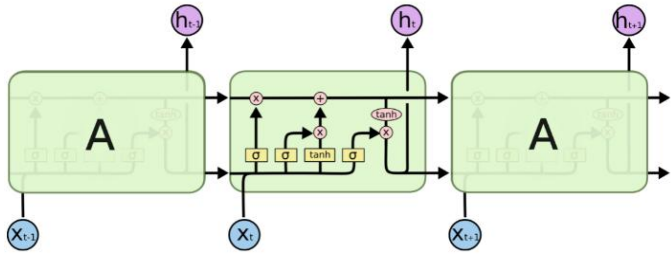

Şekil. 1. Uzun kısa-vadeli bellek[16]

TYSA tek katmandan oluşurken UKVB ise unutma kapısı katmanı, giriş kapısı katmanı ve çıkış kapısı katmanı olmak üzere üç katmandan oluşur.

Unutma kapısı bir önceki bilginin ne kadarının unutulacağı ne kadarının bir sonraki durumda kullanılmak üzere aktarılacağı kararını veren katmandır. Çıkışta 0-1 aralığında değer alır."0" eski bilgiyi tamamen unuturken "1" eski bilgiyi tamamen saklar.

$$f_t = \sigma * (W_f * [h_{t-1}, x_t] + b_f) \quad (2)$$

İkinci aşama giriş kapısı ve "tanh" katmanıdır. Bu katmanda hücre durumuna eklenecek veriler belirlenir. İlk adım $[h_{t-1}, x_t]$ değerinin giriş kapısı katmanında sigmoid fonksiyonu ile filtrelenmesidir. İkinci adım ise $[h_{t-1}, x_t]$ değerinin "tanh" katmanından geçirilmesidir. İki adımın sonucunun ve unutma kapısı katmanının sonucunun hücre durumuna eklenmesi ile yeni hücre durumu elde edilmiş olur.

$$i_t = \sigma * (W_i * [h_{t-1}, x_{t-1}] + b_i) \quad (3)$$

$$\tilde{C}_t = \tanh * (W_c * [h_{t-1}, x_{t-1}] + b_c) \quad (4)$$

$$C_t = f_t * C_{t-1} + i_t * \tilde{C}_t \quad (5)$$

Son aşama olarak da $[h_{t-1}, x_t]$ giriş değeri sigmoid fonksiyonu ile filtrelenir ve elde edilen değer hücre durumunun "tanh" katmanından geçirilmiş durumu ile çarpılır ve $h_t$ sonucu elde edilir.

$$o_t = \sigma \times (W_o \times [h_{t-1}, x_{t-1}] + b_o) \quad (6)$$

$$h_t = o_t \times \tanh(C_t) \quad (7)$$

### C. Dinamik Zaman Bükme Algoritması

İnsanlar aynı kelimeleri farklı zamanlarda farklı hızda söyleyebilmektedir. Zamandaki bu kayma problemini çözmek için Dinamik Zaman Bükme Algoritması (Doddington1971) geliştirilmiştir [8]. Eğitim ve test seti arasındaki minimum uzaklığı bulacak şekilde bir bükme fonksiyonu oluşturulur. Bu şekilde zaman bağımlılığı ortadan kaldırılarak eğitim setine en yakın test seti belirlenmiş olur.

$$d = d(E_i, T_j) = \sqrt{(E_i - T_j)^2} \quad (8)$$

$$D(i,j) = min[D(i-1,j-1), D(i-1,j), D(i,j-1)] + d \quad (9)$$

## IV. KULAK TANIMA ALGORİTMALARI

### A. DenseNet

DenseNet, evrişimsel yapay sinir ağlarının bir türüdür. Standart evrişimsel yapay sinir ağında bir görüntü sırayla birçok konvolüsyon işleminden geçerek yüksek seviyede öznitelik elde edilmektedir. Standart evrişimsel yapay sinir ağın yapısı bir katmanın çıkışı sonraki katmanın girişi şeklindedir.

DenseNet ağ yapısında ise bir katman kendisinden önceki bütün katmanların çıkışından öznitelikleri alır ve sonraki katmanlara bu bilgileri aktarır. Bu sayede her katman önceki öznitelik vektörlerine sahip olacağından ağ daha küçük olur. Bu durum, bellek boyutu ve hız açısından diğer evrişimsel yapay sinir ağlarına göre bir avantaj sağlamaktadır [11].

### B. Histogram of Gradients (HOG)

İlk olarak Dalal ve Triggs tarafından insan tespiti için önerilen bir yöntemdir. İlk adımda görüntünün gradyanları hesaplanır. Hesaplama işlemi için görüntünün renk ve yoğunluk değerleri filtrelenir. İkinci adımda görüntü küçük hücrelere bölünerek, her hücre için gradyan yönlendirmelerinin histogramları hesaplanır. Gradyanı büyük olan piksellerin histograma katkısı küçüklere göre daha fazla olmaktadır. Daha sonra komşu hücreler gruplandırılır ve ardından aydınlanma ve kontrasttaki değişikler göz önünde bulundurularak normalize edilir. Bu normalleştirme işlemi büyük bloklarda kaydırılarak gerçekleştirilir ve blokların bazı bölümleri üst üste gelebilmektedir. Bu işlemin sonunda da öznitelik vektörü elde edilmiş olur [12].

## V. SKOR SEVİYESİ FÜZYONU

Füzyon, farklı algoritmalar tarafından elde edilen sonuçları birleştirerek en iyi sonucu elde etmeyi amaçlamaktadır. Literatürde füzyon sürecinde kullanılan sensör, öznitelik, skor, sıra ve karar seviyesi olmak üzere 5 çeşit füzyon seviyesi metodolojisi bulunmaktadır [9]. Bu çalışmada skor seviyesi füzyon metodolojisi uygulanmıştır.

Skor seviyesi füzyon metodolojisi genel olarak klasik, hiyerarşik, kaskat ve hibrit olmak üzere 4 farklı skor seviyesi füzyon mekanizmasını içermektedir. Bu uygulamada hiyerarşik skor seviyesi füzyon mekanizması kullanılmış olup Şekil-2'te sunulmuştur. Hiyerarşik skor seviyesi füzyon mekanizması, çoklu algoritma ve çoklu modeli birleştirmek için ideal bir yapıya sahiptir. Füzyon 1 ve füzyon 2 çoklu algoritmayı füzyon 3 ise çoklu modeli temsil etmektedir.

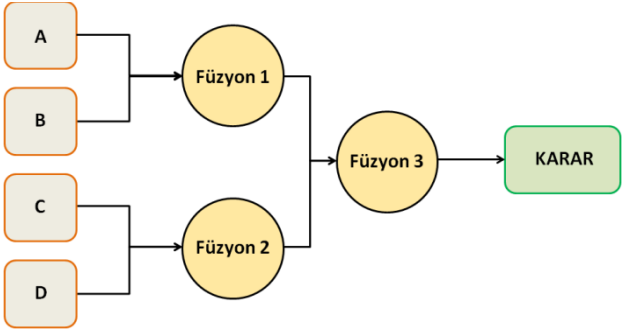

Şekil. 2.Hiyerarşik skor seviyesi füzyon mekanizması

## VI. DENEYSEL SONUÇLAR

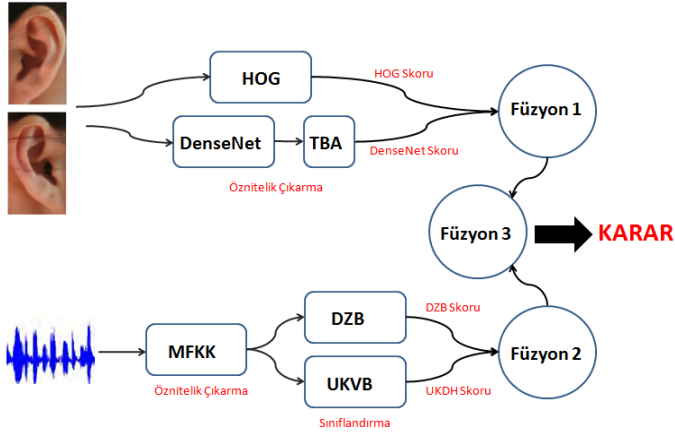

Şekil. 3.Çoklu biyometrik sistem blok diyagramı

Kulak tanımada ilk önce DenseNet ağı kullanılarak öznitelik çıkarılmıştır. Ancak, öznitelik boyutunun büyük olmasından dolayı Temel Bileşen Analizi (TBA) ile boyut küçültme işlemi yapılmıştır. İlk modaliteden elde edilen tanıma oranı %51,32'dir. İkinci modalite olarak HOG kullanılmış olup bu modalitenin tanıma oranı %38,16 olarak hesaplanmıştır. Her iki modalite için *"Manhattan"* metrik uzayı kullanılarak skor eşleştirme matrisleri elde edilmiştir. Daha sonra, çoklu-algoritma kapsamında iki modalite skor seviyesi füzyon ile birleştirilmiştir. Füzyon sonucunda %52,63 tanıma oranı bulunmuştur.

Ses tanıma için kullanılacak olan öznitelikler MFKK algoritması kullanarak elde edilmiştir. İlk modalitede DZB ile sınıflandırma yapılmıştır ve tanıma oranı %90,79 bulunmuştur. İkinci modalite olarak UKVB kullanılmış olup tanıma oranı %61,84 olarak hesaplanmıştır.

Kulak ve ses için tanıma skorları elde edildikten sonra bütün skorların aynı ölçekte olması için normalleştirme işlemi olarak *"tanh"* normalizasyonu kullanılmıştır. Bu işlem, yukarıda bahsedilen kulak ve ses için çoklu algoritma uygulamasında da gerçekleştirilmiştir. Özet olarak ilk önce çoklu algoritma daha sonra çoklu model için hiyerarşik skor seviyesi füzyon mekanizması uygulanmıştır.

İki kulak algoritmasının skor seviyesi füzyon işlemi sonucu tanıma oranı sıra-1'de %52,63, sıra-2'de %67,11 olarak elde edilmiştir.

İki ses algoritmasının skor seviyesi füzyon işlemi sonucu tanıma oranı sıra-1'de %93,42, sıra-2'de %97,37 olarak elde edilmiştir.

Kulak ve ses algoritmalarının kümülatif eşleştirme karakteristik eğrileri Şekil 4'te gösterilmiştir.

İki modalitenin çoklu algoritma sonucundan elde edilen tanıma skorlarına tekrar skor seviyesi füzyon uygulanmıştır. Bunun sonucunda tanıma oranı sıra-1'de %94,74, sıra-2'de %100 olarak elde edilmiştir.

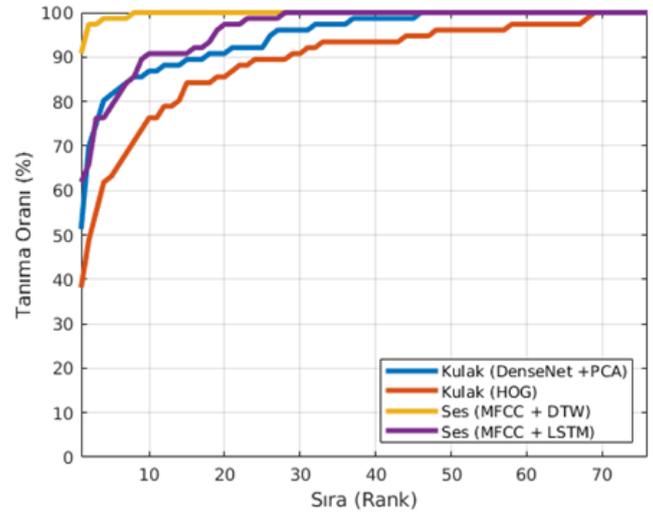

Şekil. 4.Kulak ve ses tanıma oranları

Özet olarak aşağıdaki işlemler sıralanmıştır:

- Kulak biyometrik sistemi (çoklu-algoritma)
- Ses biyometrik sistemi (çoklu -algoritma)
- Kulak ve ses biyometrik sistemi (çoklu -model)
- Hiyerarşik skor seviyesi füzyon mekanizması

Füzyon sonuçlarının kümülatif eşleştirme karakteristik eğrileri Şekil 5'te ve tüm tanıma sonuçları Tablo 1'de sunulmuştur.

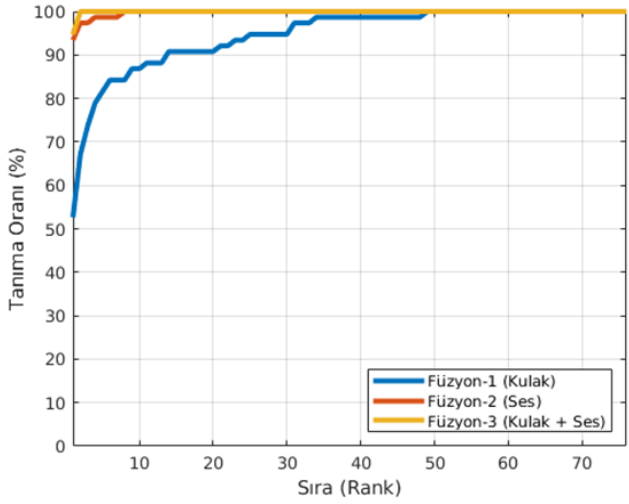

Şekil. 5. Füzyon sonuçları tanıma oranları

TABLO I. SINIFLANDIRMA SONUÇLARI

| Biyometrik Sistemler | Füzyon Ağırlığı | Sıra-1 | Sıra-2 | Sıra-5 | AUC[1] |
|---|---|---|---|---|---|
| Kulak (Gabor + DCVA) [10] | 0.15 | 43,4 | 55,3 | 71,1 | - |
| Ses (MFKK + DZB) [10] | 0.85 | 80,3 | 89,5 | 96,1 | - |
| Füzyon-1 (Kulak + Ses) [10] | - | 81,6 | 94,7 | **100** | - |
| Kulak (HOG) | 0.21 | 38,16 | 48,68 | 63,16 | 89,22 |
| Kulak (DenseNet +TBA) | 0.79 | 51.32 | 69,74 | 81,58 | 94,80 |
| Ses (MFKK + DZB) | 0.98 | 90,79 | 97,37 | 98,68 | 99,80 |
| Ses (MFKK + UKVB) | 0.02 | 61,84 | 65,79 | 78,95 | 96,31 |
| Füzyon-1 (Kulak) | 0.14 | 52,63 | 67.11 | 81,58 | 94,87 |
| Füzyon-2 (Ses) | 0.86 | 93,42 | 97,37 | 98,68 | 99,82 |
| Füzyon-3 (Kulak + Ses) | - | **94,74** | **100** | **100** | **99,97** |

## VII. SONUÇ

Daha önceki çalışmada [10] çoklu model ses ve kulak tanıma sistemi tasarlanmış olup söz konusu çalışmadaki tanıma oranı sıra-1 için %81,6, sıra-2 için %94,7'dir. %100 tanıma oranı ise sıra-5'te elde edilmiştir. Bu çalışmada kullanılan ses kayıtlarından sessiz alanlar çıkarılarak MFKK hesaplanmıştır. Bunun sonucunda tanıma oranında daha önceki çalışmaya göre %10'luk bir iyileştirme sağlanmıştır. Bu çalışmada, çoklu algoritma ve çoklu model içeren hiyerarşik skor seviyesi füzyon mekanizması kullanılarak çoklu biyometrik sistem gerçekleştirilmiş olup sıra-1'de tanıma oranı %94,74, sıra-2'de %100'dür. Çoklu model sisteme çoklu algoritma eklenerek sistemin %100 tanıma oranı sıra-2'de elde edilmiştir. Veri setimizin boyutunun küçük olması ve kontrollü olmasından dolayı tanıma oranı yüksek çıkmıştır, daha sonraki çalışmalarda kontrolsüz ortamlardan toplanan büyük veri seti ile çalışılacaktır.

---

[1] Eğri altındaki alan (area under curve)